\documentclass[letterpaper, 10 pt, conference]{ieeeconf}  

\IEEEoverridecommandlockouts                              

\usepackage[backend=biber, style=numeric, citestyle=numeric-comp, sorting=none,minbibnames=3, maxbibnames=3]{biblatex} 
\usepackage{graphics} 
\usepackage{epsfig} 
\usepackage{amsmath} 
\usepackage{amssymb}  
\usepackage{wrapfig}
\usepackage{booktabs}
\usepackage[capitalise]{cleveref}
\usepackage{siunitx}
\usepackage{dsfont}

\title{\LARGE \bf
Learning Tactile Insertion in the Real World%
}

\author{Daniel Palenicek,$^{*\dagger}$ Theo Gruner,$^{*\dagger}$ Tim Schneider,$^{*\dagger}$ Alina Böhm,$^{*\dagger}$ Janis Lenz,$^{*\dagger}$ Inga Pfenning,$^{*\dagger}$\\ Eric Krämer$^{\dagger}$ and Jan Peters$^{\dagger}$
\thanks{$^{*}$Equal contribution. $^{\dagger}$Technical University of Darmstadt, Germany.}
\thanks{This project has received funding from the Aristotle BMBF project, BMWSB ZukunftBau (no. 10.08.18.7-21.34.) and was supported by the Research Clusters “The Adaptive Mind” and “Third Wave of AI”, funded by the Excellence Program of the Hessian Ministry of Higher Education, Science, Research and the Arts, hessian.AI.}
}

\begin{document}

\maketitle
\thispagestyle{empty}
\pagestyle{empty}

\begin{abstract}
    Humans have exceptional tactile sensing capabilities, which they can leverage to solve challenging, partially observable tasks that cannot be solved from visual observation alone.
    Research in tactile sensing attempts to unlock this new input modality for robots.
    Lately, these sensors have become cheaper and, thus, widely available.
    But, how to integrate them into control loops is still an active area of research, with central challenges being partial observability and the contact-rich nature of manipulation tasks.
    In this study, we propose to use Reinforcement Learning to learn an end-to-end policy, mapping directly from tactile sensor readings to actions.
    Specifically, we use Dreamer-v3 on a challenging, partially observable robotic insertion task with a Franka Research 3, both in simulation and on a real system.
    For the real setup, we built a robotic platform capable of resetting itself fully autonomously, allowing for extensive training runs without human supervision.
    Our initial results show that Dreamer is capable of utilizing tactile inputs to solve robotic manipulation tasks in simulation and reality.
    Further, we find that providing the robot with tactile feedback generally improves task performance, though, in our setup, we do not yet include other sensing modalities.
    In the future, we plan to utilize our platform to evaluate a wide range of other Reinforcement Learning algorithms on tactile tasks.
\end{abstract}

\section{Introduction}

Humans heavily rely on their tactile sense to solve dexterous manipulation tasks~\cite{klatzky1995identifying, johansson2009coding}. 
Haptic feedback allows humans to achieve high levels of accuracy in tasks ranging from classical assembly tasks to critical medical surgeries.
At the same time, the dexterous manipulation capabilities of even the most sophisticated robots lag behind those of even a small child if the environment is not strictly controlled.
One way of closing this gap is to equip robots with tactile sensors.
Tactile sensors provide robots with crucial feedback at the points of contact, which their end-effectors often occlude from vision.

Recently, the development of vision-based tactile sensors~\cite{yuan2017gelsight,ward2018tactip,lambeta2020digit} has sparked interest in applying machine learning to various tactile-based perception tasks, such as object classification~\cite{Corradi2015May}, texture recognition~\cite{böhm2024matters,Yuan2018May}, and shape reconstruction~\cite{suresh2022shapemap}. 
In contrast to conventional sensors that estimate physical quantities like contact forces or torques, vision-based tactile sensors provide an image of the deformation of the contact surface. 
This shift in paradigm necessitates pre-processing the sensor data before utilization in a control loop.
However, advances in computer vision in the last decades have made machine learning methods effective in extracting information from visual representations~\cite{lecun1989handwritten,he2015deep}.

\begin{figure}[t]
    \centering
    \includegraphics[width=\linewidth]{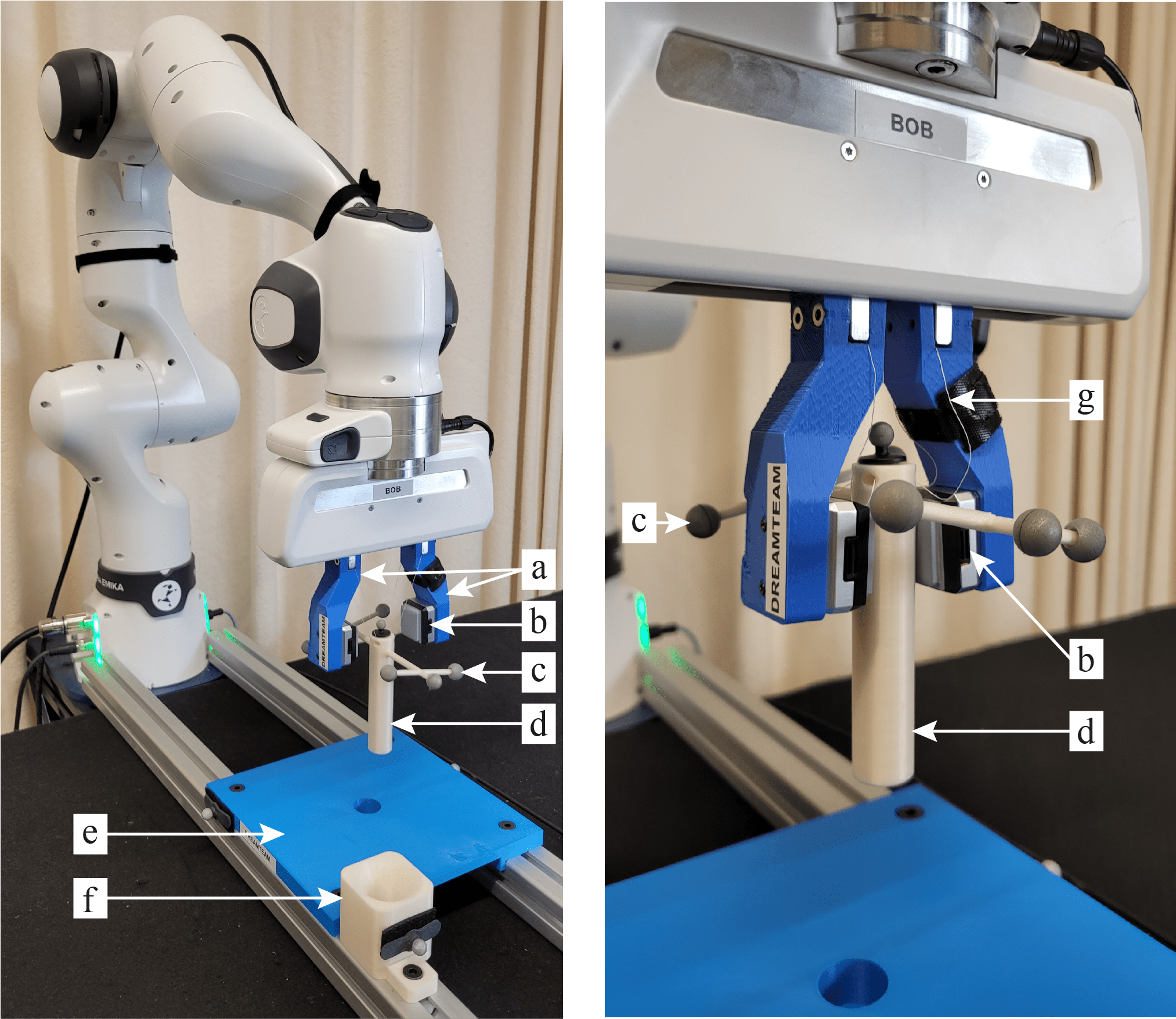}
    \caption{%
        \textbf{Tactile Insertion Platform Setup.} We label the gripper (a), GelSight Mini sensor (b), OptiTrack marker (c), cylinder (d), base plate (e), reset box (f), and thin thread (g).
        Note, that only one of the two GelSight Mini sensors is used in this work.
    }
    \label{fig:setup_real}
    \vspace{-1.5em}
\end{figure}

A central challenge of many manipulation tasks is partial observability, as object tracking is often inaccurate, especially due to occlusion caused by the robot's end-effector.
While tactile sensors can provide crucial information about the points of contact, the information obtained from them will usually still be incomplete, rendering the decision-making problem a \emph{Partially Observable Markov Decision Process (POMDP)}.
Yet, reinforcement learning (RL) algorithms and model-based RL algorithms commonly rely on the assumption that the state-space~$\mathcal{S}$ is Markovian and fully observable~\cite{schulman2017ppo,haarnoja2018sac,fujimoto2018td3,schneider2022active,palenicek2023diminishing,bhatt2023crossq}.
Recently, a number of RL algorithms have been proposed that tackle POMDPs by learning a latent state-space model of the environment~\cite{Hafner_2019_LatDynPixels,lee2020stochastic,Havens_Ouyang_LearningLatentStateSpaces,Rybkin_2021_MBRLViaLatentSpaceColloc,Nguyen_2021_TemporalPredictiveCodingForMB,Gelada_et_al_2019,Hafner_2020_DreamToControl,Hafner_2020_MasteringAtari,Wu_hafner_2022_DayDreamer}.

In this paper, we utilize \emph{Dreamer-v3}~\cite{hafner2023mastering} to solve a challenging, partially observable robotic insertion task purely from tactile feedback, both in simulation and the real world.
Dreamer is a model-based RL approach that learns a latent state-space model and an actor-critic model simultaneously, which has shown great success in challenging Atari games~\cite{Hafner_2020_MasteringAtari} and even on a physical quadruped task~\cite{Wu_hafner_2022_DayDreamer}.
We show that Dreamer is capable of utilizing tactile feedback from a \emph{Gelsight Mini}~\cite{yuan2017gelsight} sensor effectively while simultaneously being sample efficient enough to train in the real world from scratch.
Our contributions are (i) the development of a tactile insertion simulation setup based on \emph{Taxim}~\cite{taxim}, allowing for rapid development and testing, (ii) the design of a real robot platform with fully autonomous robust resetting of the task, allowing for extensive training runs without human supervision, and (iii) a preliminary evaluation the influence of touch on learning insertion policies in simulation and on the physical system. 

\section{Robotic Platform}

\begin{figure}[t]
    \centering
    \includegraphics[width=\linewidth]{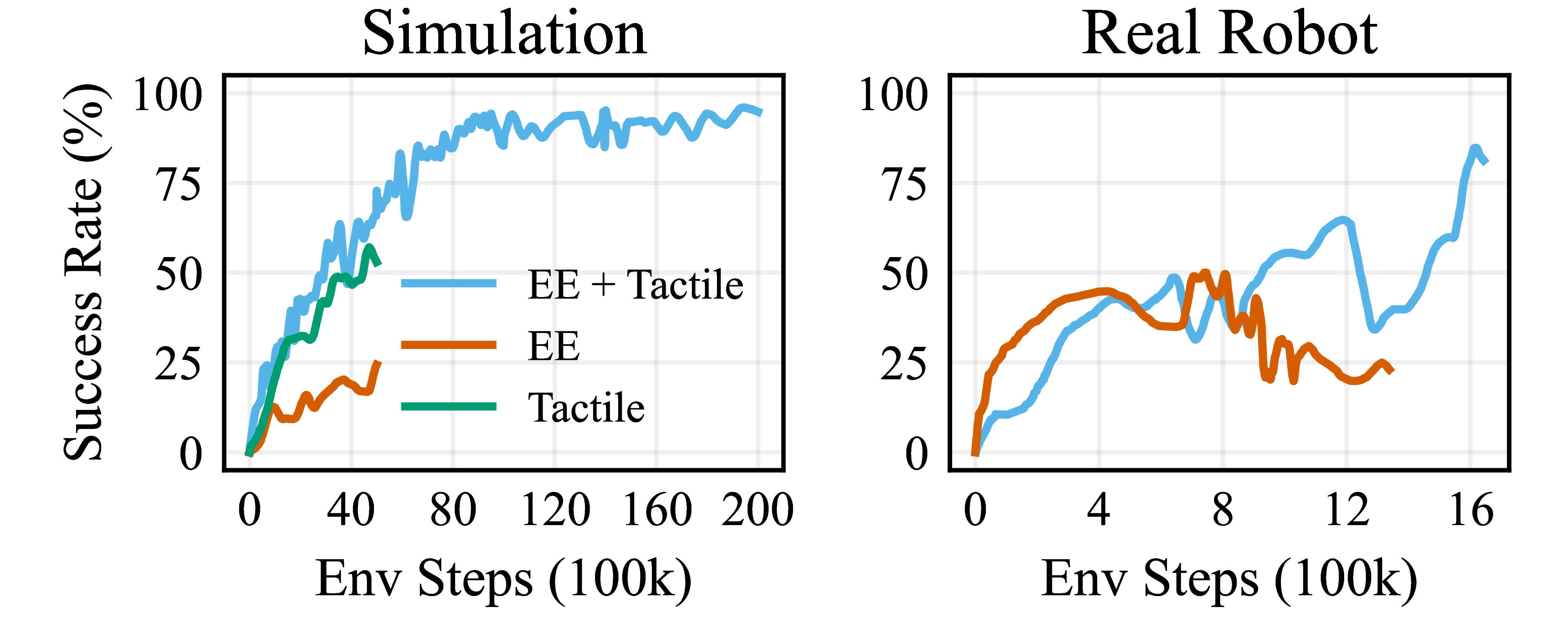}
    \vspace{-1.5em}
    \caption{%
        \textbf{Evaluation Results.}
        Insertion success rates during training in simulation (left) and on the real system (right).%
    }
    \vspace{-1.5em}
    \label{fig:evaluation}
\end{figure}

The robotic setup, consisting of a 7 DoF Franka Research~3, a circular peg, and a base plate with an insertion hole, is presented in~\ref{fig:setup_real}.
We mount a vision-based GelSight Mini tactile sensor~\cite{yuan2017gelsight} sensor in the fingertips of the robot's gripper. 
A key challenge is the development of a reset procedure that places the agent back in its initial state, with the peg properly placed in its gripper.
To prevent the loss of the cylinder if the peg slips from the gripper, we suspend the peg with a thin thread connected to the gripper. Upon losing contact, the peg hangs from the thread. 
The robot then moves to a reset box, lowers the peg inside, re-grasps, and returns to its initial position above the insertion hole. 
This resetting procedure enables the agent to train consecutively for multiple days without requiring human intervention.

The agent's observation comprises a $\numproduct{64x64x3}$ RGB image from the Gelsight sensor and the end-effector position $\mathbf{p}_e\in\mathbb{R}^3$ in Cartesian coordinates at \SI{25}{\hertz}. 
The policy returns the relative new positions of the end-effector at \SI{20}{\hertz}.
These target positions are then fed into the \emph{franky}~\cite{franky} control library, which internally uses \emph{ruckig}~\cite{berscheid2021jerk} for smooth motion planning and Franka's internal cartesian impedance controller for execution. 
We ensure safe exploration of the environment by restricting the workspace $\mathcal{W}$ to be in the vicinity of the hole.
For continuous tracking of the cylinder pose, we utilize OptiTrack.
This information is only used for evaluation of the robot's performance and not as an input to the policy.
\begin{wraptable}{r}{42mm}
    \vspace{-1.5em}
    \centering
    \caption{\textbf{Reward function,} with multiple components.}
    \begin{tabular}{l}
       \toprule
       $r \hspace{.4em} = r_d + r_g + r_p + r_{\mathbf{a}}$ \\
       \midrule
       $r_d =-5\cdot (|\mathbf{p}_g - \mathbf{p}_e|)$ \\
       $r_g = 100\cdot \mathds{1}_{\{\mathcal{G}\}}(\mathbf{p}_g)$\\
       $r_{p} = -100 \cdot \mathds{1}_{\{\mathbb{R}^3 / \mathcal{W}\}}(\mathbf{p}_e)$\\
       $r_{\mathbf{a}} = \num{e-3} \cdot |\mathbf{a}|$\\
       \bottomrule
    \end{tabular}
    \label{table:reward}
    \vspace{-1.5em}
\end{wraptable}
To simulate variation of the hole position, we add a random, per episode constant, offset of up to \SI{5}{\milli\m} to the observed end-effector position.
This way, even though the hole position never changes, for the agent it seems as if it did since its base-frame is slightly different in every episode.
Furthermore, we randomly vary the starting position of the robot's end-effector. 

As shown in \cref{table:reward}, the reward comprises of four components: (i) proximity to the goal, (ii) a terminal reward upon reaching the goal $\mathcal{G} = \{\mathbf{x}\in\mathbb{R}^3: |\mathbf{p}_g - \mathbf{x}| < (5, 5, 10)[\si{\milli\metre}]\}$, (iii) a terminal penalty for leaving the workspace, and (iv) an action penalty to encourage smooth motions. 
The policy training includes several neural network models: encoder and decoder networks that project between the observation space and the 4160-dimensional latent space, a recurrent state-space model, a reward model in the latent space, and the policy.
In addition, we develop a digital twin of our setup in PyBullet~\cite{coumans2016pybullet}, tactile sensors are simulated with Taxim~\cite{taxim}.

\section{Empirical Evaluations}

We train Dreamer from scratch, both in simulation and reality.
We conduct experiments with and without tactile sensors to evaluate the benefits of tactile sensing.

\paragraph{\textbf{Simulation results}}
\cref{fig:evaluation} shows that Dreamer achieves a success rate up to $90\%$ given end-effector and tactile information.
We investigate the importance of proprioception and tactile sensing by removing the end-effector and tactile observations in two experiments. 
The results clearly show that using tactile images greatly improves learning performance, while observing the end-effector position yields no significant benefit.

\paragraph{\textbf{Real Robot results}}
On the real setup, our study comprises an experiment using only the end-effector position and one including the tactile feedback.
Although the difference in performance between these two experiments is less pronounced than in simulation, especially in the beginning, tactile sensing still seems to provide a substantial advantage in performance.
We hypothesize that the inherent softness of the gripper in the real system might simplify the task compared to the simulated setup, where we use rigid body physics.
However, the results are still preliminary, and more experiments will have to be conducted in the future.

\section{Conclusion \& Future Work}

In this work, we developed a robot platform that facilitates autonomous training of an RL agent on a Franka Research~3 robot to investigate whether RL is a viable option for incorporating tactile feedback in control loops. 
To ensure reliable and safe training, we implement a robust resetting routine and restrict the action space, enabling autonomous learning of an insertion policy without human supervision. 
Leveraging tactile feedback from a vision-based tactile sensor attached to the gripper, our policy autonomously learns to insert a peg into a hole using the model-based RL approach Dreamer.
Our simulation results suggest that the inclusion of tactile information significantly enhances learning outcomes. 
With our platform, we intend to benchmark other RL algorithms on tactile insertion tasks to get a clear understanding of which methods are capable of dealing with the inherent complexity of tactile manipulation tasks in the real world.
Furthermore, we plan to increase task complexity in future experiments, enabling a thorough exploration of touch's role in dexterous manipulation.

\printbibliography
\clearpage

\end{document}